\pdfoutput=1
\documentclass[lettersize,journal]{IEEEtran}
\usepackage{etoolbox}
\usepackage{booktabs}
\usepackage{cite}
\usepackage{amsmath,amssymb,amsfonts}
\usepackage{graphicx}
\usepackage{textcomp}
\usepackage{float}
\usepackage{multirow}
\usepackage{makecell}
\usepackage{autobreak}
\usepackage{balance}
\usepackage{hyperref}
\usepackage[ruled]{algorithm2e}
\usepackage{url} 
\usepackage{footnote}
\usepackage{bbding}
\usepackage{cleveref}
\Crefname{figure}{Fig.}{Figs.}
\Crefname{table}{Table.}{Tables.}
\Crefname{equation}{Equation.}{Equations.}

\def\BibTeX{{\rm B\kern-.05em{\sc i\kern-.025em b}\kern-.08em
    T\kern-.1667em\lower.7ex\hbox{E}\kern-.125emX}}
\begin{document}
\title{X-Recon: Learning-based Patient-specific High-Resolution CT Reconstruction from Orthogonal X-Ray Images}
\author{Yunpeng Wang$^{\dagger}$, Kang Wang$^{\dagger}$, Yaoyao Zhuo, Weiya Shi, Fei Shan$^{\ast}$, and Lei Liu$^{\ast}$
\thanks{Manuscript received December 31, 2023. This work was supported by the Peak Disciplines (Type IV) of Institutions of Higher Learning in Shanghai. (YP. Wang and K. Wang contributed equally to this work.)(Corresponding authors: Fei Shan and Lei Liu.)}
\thanks{YP. Wang is with the Institutes of Biomedical Sciences, Fudan University, Shanghai 200032, China (e-mail:wangyp18@fudan.edu.cn).}
\thanks{K. Wang is with the School of Basic Medical Sciences, Fudan University, Shanghai 200032, China. (e-mail:wangk18@fudan.edu.cn)}
\thanks{YY. Zhuo is with the Department of Radiology, Zhongshan Hospital, Fudan University, Shanghai 200032, China, and is also with the Shanghai Institute of Medical Imaging, Shanghai 200032, China. (e-mail:zhuo.yaoyao@zs-hospital.sh.cn)}
\thanks{WY. Shi and F. Shan are with the Department of Radiology, Shanghai Public Health Clinical Center, Fudan University, Shanghai 200433, China. (e-mail:shiweiya@shaphc.org, shanfei\_2901@163.com)}
\thanks{L. Liu is with the Intelligent Medicine Institute, Shanghai Medical College, Fudan University, Shanghai 200032, China, and is also with the Shanghai Institute of Stem Cell Research and Clinical Translation, Shanghai 200123, China. (e-mail:liulei\_sibs@163.com)}
}
\maketitle

\begin{abstract}
    Rapid and accurate diagnosis of pneumothorax, utilizing chest X-ray and computed tomography (CT), is crucial for assisted diagnosis. Chest X-ray is commonly used for initial localization of pneumothorax, while CT ensures accurate quantification. However, CT scans involve high radiation doses and can be costly. To achieve precise quantitative diagnosis while minimizing radiation exposure, we proposed X-Recon, a CT ultra-sparse reconstruction network based on ortho-lateral chest X-ray images. X-Recon integrates generative adversarial networks (GANs), including a generator with a multi-scale fusion rendering module and a discriminator enhanced by 3D coordinate convolutional layers, designed to facilitate CT reconstruction. To improve precision, a projective spatial transformer is utilized to incorporate multi-angle projection loss. Additionally, we proposed PTX-Seg, a zero-shot pneumothorax segmentation algorithm, combining image processing techniques with deep-learning models for the segmentation of air-accumulated regions and lung structures. Experiments on a large-scale dataset demonstrate its superiority over existing approaches. X-Recon achieved a significantly higher reconstruction resolution with a higher average spatial resolution and a lower average slice thickness. The reconstruction metrics achieved state-of-the-art performance in terms of serveral metrics including peak signal-to-noise ratio. The zero-shot segmentation algorithm, PTX-Seg, also demonstrated high segmentation precision for the air-accumulated region, the left lung, and the right lung. Moreover, the consistency analysis for the pneumothorax chest occupancy ratio between reconstructed CT and original CT obtained a high correlation coefficient. Code will be available at: https://github.com/wangyunpengbio/X-Recon
\end{abstract}


\section{Introduction}
\label{sec:introduction}
\IEEEPARstart{C}{linically}, an abnormal accumulation of air in the pleural cavity between the lungs and the chest wall is referred to as pneumothorax~\cite{bintcliffe_spontaneous_2014}. Pneumothorax is a serious disease and a relatively common clinical problem, with an annual incidence measured at 22.7 cases per 100,000 people~\cite{bobbio_epidemiology_2015}. Patients with pneumothorax may experience symptoms such as chest pain, lung atrophy, dyspnea, and, in severe cases, shock and asphyxia~\cite{bobbio_epidemiology_2015}. In clinical practice, X-rays and computed tomography (CT) scans are commonly used to diagnose pneumothorax~\cite{urooj_computer-aided_2022}. Specifically, the two-dimensional (2D) radiography (X-ray) is less radiation-intensive. Nearly all hospitals, including those in remote and technologically underserved rural areas, are equipped with X-ray machines. These machines provide X-ray images for patients, facilitating primary care. In contrast, three-dimensional (3D) CT provides more detailed images of the body's interior, which can be used to diagnose the presence of a pneumothorax and estimate its size and location, ascertain if it's a tension pneumothorax, and evaluate for mediastinal displacement. Nevertheless, CT equipment is expensive, and CT scans involve relatively higher radiation exposure for the long scanning time. Consequently, it's of clinical significance to devise techniques for producing high-quality CT images with reduced radiation exposure.

Driven by the above requirements and inspired by X-ray and CT acquisition principles, current research has attempted to generate high-precision 3D CT images using low-dose biplane X-ray images. This way of modeling offers two significant advantages. Firstly, it allows for a substantial reduction in the radiation dose absorbed by patients, decreasing the radiation dose range from 1 to 10 mSv for chest CT images to as low as 0.02 mSv when using biplane X-ray images~\cite{mccollough_answers_nodate,atli_radiation_2021}. Additionally, these methods only necessitate basic biplane X-ray images, and highly efficient approaches, such as our proposed method, have the potential to be implemented on existing X-ray machines without incurring extra equipment expenses. This could have a positive impact on a larger number of patients, particularly in developing countries.

Generally, the process of projecting complex 3D geometries into flat 2D images introduces information loss. Therefore, reconstructing a 3D image from 2D images is a formidable challenge and has been a long-standing problem in the field of medical imaging as well as computer vision. These methods reconstruct the shape of 3D objects based on indirect information such as light source, focus, texture, or motion~\cite{zhou_handbook_2019}. However, traditional methods require multiple views of identical objects using a calibrated camera, which is challenging in realistic scenarios. Recent learning-based view synthesis techniques have shown more promising results in the computer vision and graphics community. A series of neural networks, represented by Nerf~\cite{mildenhall_nerf_2021}, learn robust reconstruction functions in a data-driven manner that can be adapted to perform image reconstruction in complex scenes. This has also stimulated enthusiasm in both the academic and industrial communities.

Similarly, CT reconstruction was originally realized through mathematical inversion of the projected images from multi-view images. However, traditional reconstruction algorithms require a large number of projected images to avoid artifacts, which increases exposure time and radiation dose. Although efforts have been made to reduce the number of samples used for CT reconstruction, these sparse-sampling CT reconstruction methods still require dozens of images from different shot angles and are ineffective in reducing radiation exposure. The problem of ultra-sparse reconstruction using only biplane X-ray to reconstruct CT scans has not been solved properly. In recent years, a few successful attempts at inverse mapping have emerged. Henzler et al.~\cite{henzler_singleimage_nodate} first applied a deep Convolutional Neural Network (CNN) to single-radiograph tomography and reconstructed 3D cranial volumes from 2D X-rays. Kasten et al.~\cite{10.1007/978-3-030-61598-7_12} used an end-to-end CNN for 3D reconstruction of knee bones from bi-planar X-ray images. Shen et al.~\cite{shen_patient-specific_2019} developed a deep network system with representation, transformation, and generation modules to generate volumetric tomography images from single or multiple 2D X-rays.

\begin{figure}[bp]
    \vspace{-0.4cm}
    \centering
    \includegraphics[width=\linewidth]{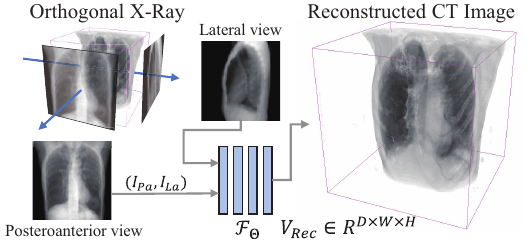}
    \caption{Patient-specific high-resolution CT reconstruction from orthogonal X-rays. $I_{Pa}$ and $I_{La}$ represent posteroanterior and lateral X-ray images and $V_{Rec}$ represents the volumetric CT image reconstructed by the network $\mathcal{F}_{\Theta}$.}
    \label{fig:intro}
    \vspace{-0.4cm}
\end{figure}

In this study, we proposed a learning-based CT reconstruction network called X-Recon for ultra-sparse 3D tomographic image reconstruction using biplane X-rays acquired from orthogonal views as shown in \Cref{fig:intro}. X-Recon utilizes a generative adversarial framework, which includes a generator equipped with a multi-scale fusion rendering module (MFusionRen) and a 3D coordinate convolution layer that replaces traditional convolution layers in the discriminator. To supervise the model, we introduce a multi-angle projection loss based on the Projective Spatial Transformer (ProST). This approach allows the incorporation of anatomical prior knowledge and the learning of the mapping relationship from X-rays to CT using a large training dataset. Extensive experiments conducted on a large-scale dataset demonstrate that X-Recon achieves a significantly higher CT reconstruction resolution, with dimensions of 224 pixels $\times$ 224 pixels $\times$ 224 pixels and an average spatial resolution of 1.6 mm. This performance sets a new state-of-the-art (SOTA) standard in ultra-sparse 3D tomographic image reconstruction, offering unprecedented resolution. For reliability evaluation, we employed a zero-shot segmentation framework called PTX-Seg, which combines traditional image processing and deep learning models to delineate the air-accumulated region and lungs for the calculation of the percentage of pleural cavity occupied by pneumothorax. The Pearson's correlation coefficient calculated between the X-Recon reconstructed CT and the real CT for pneumothorax pleural cavity occupancy ratio reached 0.77, which demonstrates its potential for clinical applicability. In summary, this study makes the following contributions:
\begin{itemize}
    \item Proposing a novel learning-based 2D-3D CT reconstruction network called X-Recon, which achieves ultra-sparse 3D tomographic image utilizing only two projection views and sets a new state-of-the-art (SOTA) standard in ultra-sparse 3D tomographic image reconstruction. 
    \item Compared with other reconstruction algorithms, X-Recon significantly elevates the resolution of reconstructed CT images to a higher level through the introduction of the MFusionRen module, coordinate convolution and multi-angle projection loss.
    \item A zero-shot pneumothorax segmentation pipeline PTX-Seg is proposed combining image processing techniques with deep learning models and used to evaluate image quality of reconstructed CT.
\end{itemize}

\section{Related Works}
\subsection{2D–3D vision reconstruction}
A series of shape-from-X methods~\cite{zhou_handbook_2019} have been proposed to recover 3D information from 2D images in the field of multi-view stereo vision. Starting from a geometrical viewpoint by providing a mathematically formulated representation of the projection process from 3D to 2D, image reconstruction is subsequently performed by dedicated algorithms to solve the inverse problem of projection. For example, binocular vision-based approaches~\cite{hartley_multiple_2003} require feature matching of different viewpoints, followed by recovering the 3D coordinates of the image pixels based on the principle of optical triangulation, whereas contour-based 3D reconstruction methods~\cite{hartley_multiple_2003} require precisely segmented 2D contours.

Recently, researchers~\cite{shen_patient-specific_2019} have been motivated by the human perception pattern, which suggests that one can infer the approximate size and geometry of an object using only one eye based on prior knowledge. Deep learning view synthesis techniques, such as Neural Radiance Fields (NeRF)~\cite{mildenhall_nerf_2021}, have the ability to directly recover the 3D geometry of an object from one or more RGB images without requiring a complex camera calibration process. These techniques transform the 3D reconstruction problem into a generative problem and incorporate prior knowledge.
NeRF~\cite{mildenhall_nerf_2021} along with its variants~\cite{mildenhall_nerf_2021, pumarola_d-nerf_nodate, xu_h-nerf_2021, gao_nerf_2022}, have shown even more promising
\begin{figure*}[htbp]
    \vspace{-0.8cm}
    \centering
    \includegraphics[width=\linewidth]{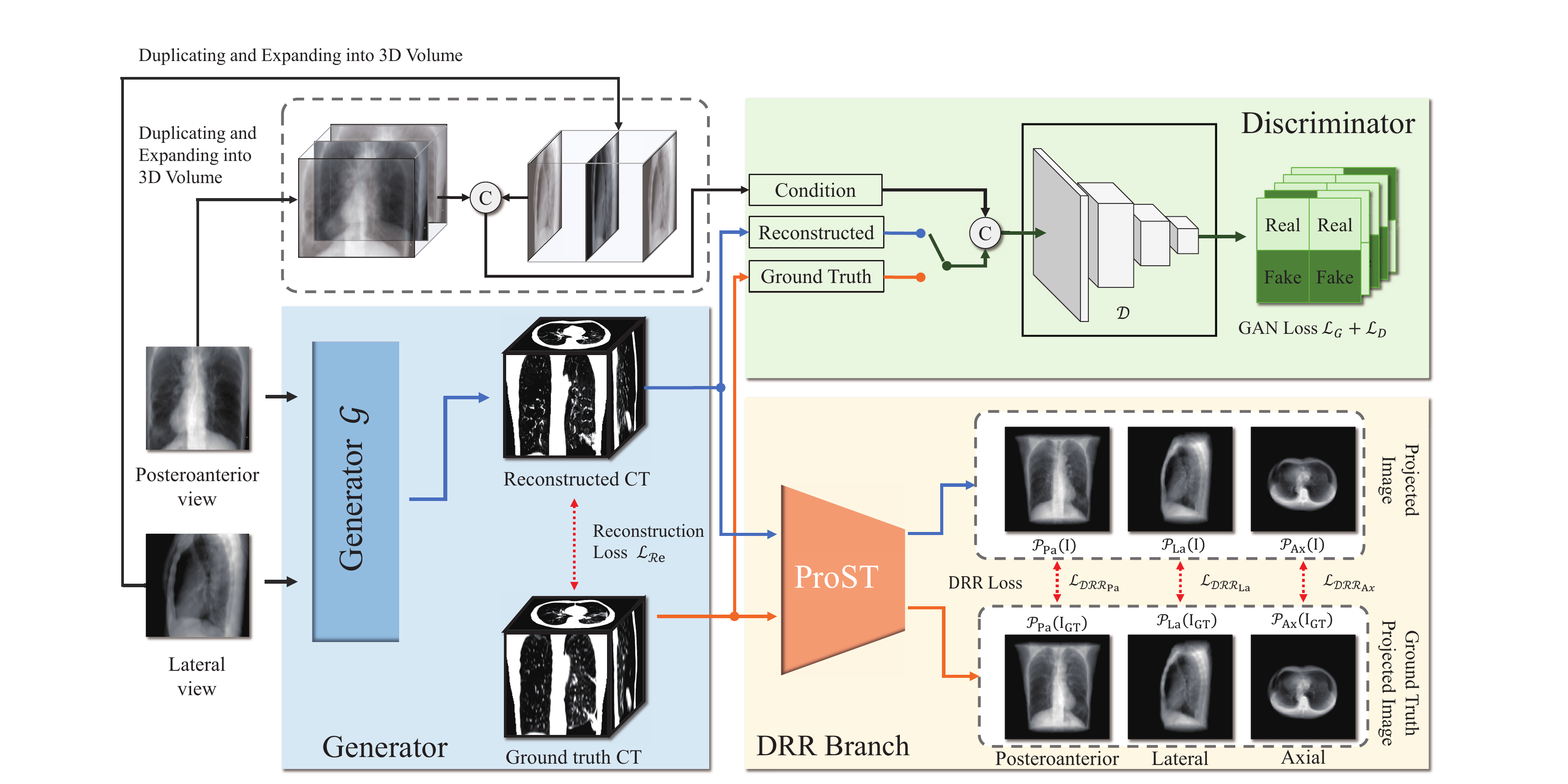}
    \caption{The overview architecture of X-Recon, the model takes posteroanterior and lateral X-ray images as inputs, and the generator produces the reconstructed CT image. The loss functions employed in the training process involve three main components: the reconstruction loss, the multi-angle digitally reconstructed radiograph loss, and the generative adversarial loss. \textcircled{\scriptsize C} denotes concatenation, straight arrows with different colors represent distinct data flows, and red dashed arrows indicate the computation of corresponding loss functions between predictions and GTs.}
    \label{fig:X-Recon_Overview}
    \vspace{-0.4cm}
\end{figure*}
results in the computer vision and graphics community. There are significant differences in the imaging principles between the reflection-based visible light imaging of such methods and the transmission-based X-ray imaging required for this study, and it is not possible to directly apply them to this study. However, the idea of introducing prior knowledge through deep learning has profoundly influenced the present work and is reflected in the study of biplane X-ray image reconstruction for CT.

\subsection{Reconstruction of volumetric images from X-rays}
In medical imaging, CT images are obtained by mathematically inverting a series of projected images from different angles. However, in traditional reconstruction algorithms, such as the filtered back projection (FBP) method, the number of projections must satisfy the Shannon/Nyquist sampling theorem to avoid streak artifacts~\cite{kharfi_mathematics_2013}, which puts an extremely high demand on the exposure time and thus increases the radiation dose. To alleviate this problem, researchers have proposed sparsely sampled CT reconstruction based on compressed perception~\cite{chen_prior_2008, yu_compressed_2009, choi_compressed_2010, zhu_ct_nodate} and maximum a posteriori~\cite{fessler_spatial_1996} methods. Such methods reduce artifacts by incorporating regularization terms during image reconstruction~\cite{stayman_regularization_2000, wang_penalized_2006, xu_low-dose_2012}. However, such sparse-sampling CT reconstruction methods still project images from at least dozens of viewpoints without significantly reducing radiation exposure if the image quality is guaranteed to be unaffected. In fact, although researchers have been trying to reduce the number of samples in CT imaging, the CT reconstruction with ultra-sparse sampling required for biplane X-ray in this study has still not been well achieved.

\section{METHODOLOGY}
In this paper, we have made the following contributions in terms of methodology: 1) a generalized CT reconstruction network is proposed for X-ray images, called X-Recon, along with several dedicated loss functions to enhance the algorithm's performance.  2) a zero-shot pneumothorax segmentation algorithm, referred to as PTX-Seg, which combines traditional image processing techniques with deep learning models, is introduced for the subsequent segmentation task. The detailed introductions and training pipelines are described in the following sections.

\subsection{X-Recon: Dual View CT Reconstruction Network}
The overall structure of X-Recon is depicted in \Cref{fig:X-Recon_Overview}, and the X-Recon follows the adversarial network paradigm with an additional digitally reconstructed radiograph (DRR) branch. In this study, the ortho-lateral chest X-ray images are simultaneously fed into the generator of X-Recon as a prior condition to obtain the reconstructed CT, which is then used to calculate the reconstruction loss between the reconstructed CTs and the CTs acquired from real patients. Additionally, the projective spatial transformer (ProST) is employed to output the reconstructed CT and employ a multi-angle projection loss between the reconstructed CT and the real CT for supervision, further enhancing the reconstructed image. Furthermore, the ortho-lateral chest X-ray image is also supplied to the discriminator as a prior condition, enhancing the information available to the discriminator and improving its performance. This, in turn, facilitates the adversarial training procedure of the generator, leading to improved image quality in the reconstructed CT.

\begin{figure*}[htbp]
    \vspace{-0.6 cm}
    \centering
    \includegraphics[width=\linewidth]{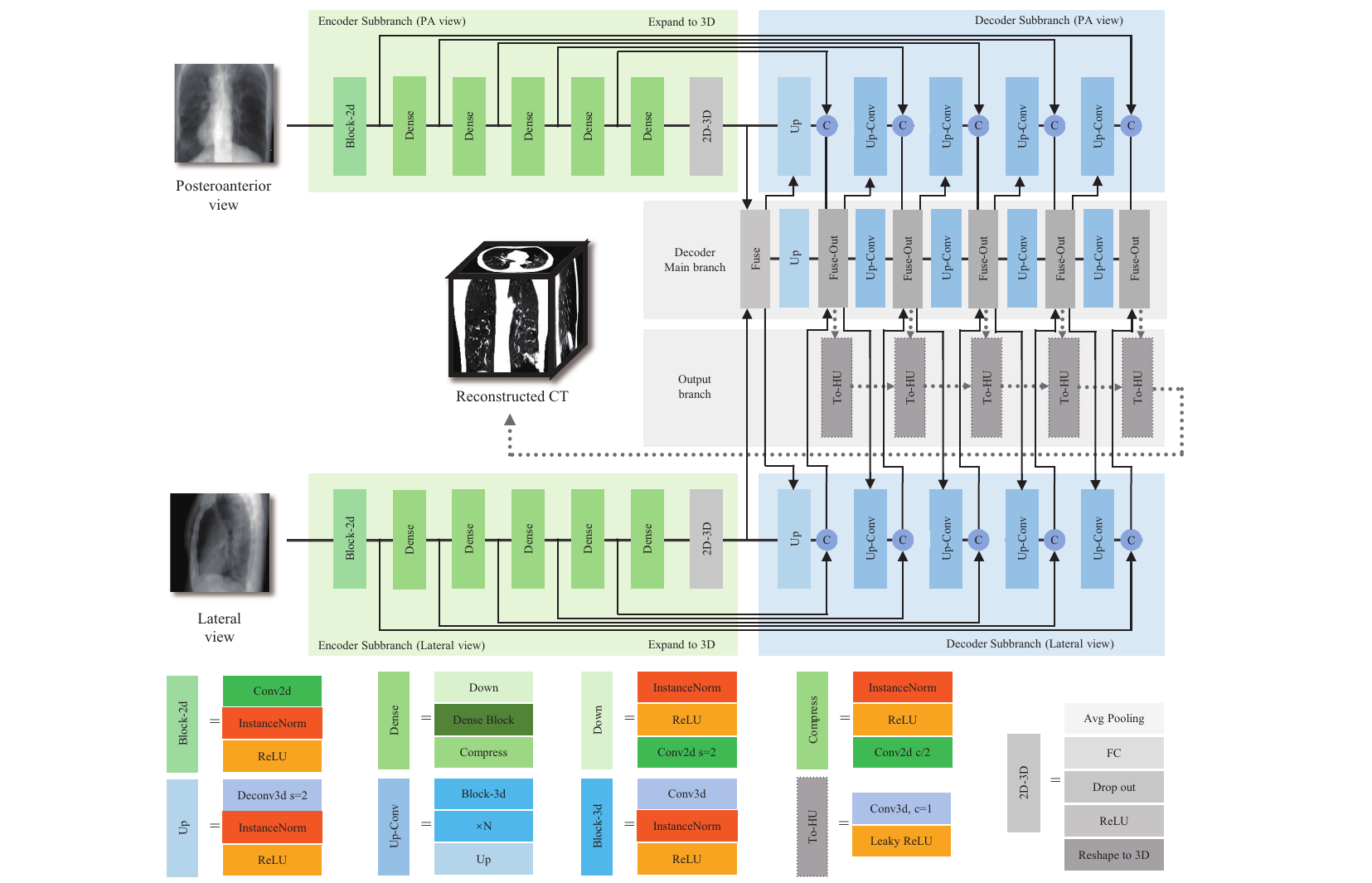}
    \caption{Details of the network structure of the X-Recon generator. It consists of two encoder-decoder subbranches as well as a decoder main branch and an output branch. Each encoder-decoder branch is used to process information from an independent viewpoint. The decoder main branch is used for information fusion and the output branch is used for CT overlay rendering.}
    \label{fig:generator}
    \vspace{-0.4 cm}
\end{figure*}

\subsubsection{Generator}\label{sec:generator}
The detailed architecture of the generator in X-Recon is illustrated in \Cref{fig:generator}, comprising three independent modules. The first two encoder-decoder components utilize the same network structure to process images from different viewpoints. The last component is the multi-scale fusion rendering (MFusionRend) module, which serves as the final prediction of the reconstructed image. Specifically, the twin encoder-decoders process posteroanterior and lateral chest X-rays in separate pathways. Because a single posteroanterior chest X-ray image cannot capture the lateral information of an object, and a lateral-only chest X-ray loses the frontal information, X-Recon addresses this limitation by taking dual orthogonal view images, including both frontal and lateral chest X-ray images, as input. The twin encoder-decoder networks extract features from these two views and forward them to the subsequent fusion and decoding module. The MFusionRend module includes a decoder main branch for integrating information from dual orthogonal viewpoints and an output branch for generating features at multiple spatial levels. 

X-Recon employs dense connection modules as the fundamental unit in the encoder. Each of these densely connected modules is composed of a down-sampling module with a step size of 2, a densely connected convolution block, and a compression block that reduces the output channel by half. The ultimate goal of these designs is to create compressed and extracted feature representations while preserving information from the input image at various spatial levels. This is achieved by cascading densely connected modules and transmitting them to the corresponding decoders through skip connections.

\subsubsection{Discriminator}: 
Unlike traditional generative adversarial networks (GANs), X-Recon constructs the discriminator with a fully convolutional form. In the traditional GAN, the discriminator outputs a single number indicating the likelihood that the input sample is real. However, in our fully convolutional discriminator network, the input image is transformed into a matrix. Each value in this matrix represents the probability that the corresponding region in the original image is real. This scheme has proved effective and previous work~\cite{isola_image--image_nodate} has demonstrated that this approach exhibits strong generalization in the domain of high-resolution and high-definition image generation. Besides, the translation invariance offered by traditional convolution is conducive to the improved learning of robust features, particularly in tasks like image classification. Nevertheless, in cases where the object within the image remains relatively stationary, the translation invariance of traditional convolution can potentially limit its capacity to capture positional information. To mitigate this problem, the Coordinate Convolution (CoordConv)~\cite{liu_intriguing_2018} was introduced. CoordConv integrates coordinate information as part of the feature map, allowing the network to maintain a degree of translational dependency during the task's learning process. In the context of CT reconstruction, the position of each organ in the chest image remains relatively fixed. Allowing the network to retain positional information for each organ can enhance the model's performance. 

With the same expectations, X-Recon extended the coordinate convolution layer to a 3D format and replaced the conventional convolution layer in the original discriminator network. Furthermore, the discriminator of X-Recon is constructed in a 3D format. Firstly, the feature extraction as well as the fusion of positional information is performed through a 3D coordinate convolutional layer. Then, the feature map is passed through three cascaded convolutional down-sampling modules. Each module consists of a 3D convolutional layer with a step of two, a normalization layer, and a corrected linear unit with leakage. Finally, the feature map is compressed by an output convolutional layer to obtain the final output matrix.

\subsubsection{MFusionRend Module}
\begin{figure}[htbp]
    \centering
    \vspace{-0.4 cm}
    \includegraphics[width=\linewidth]{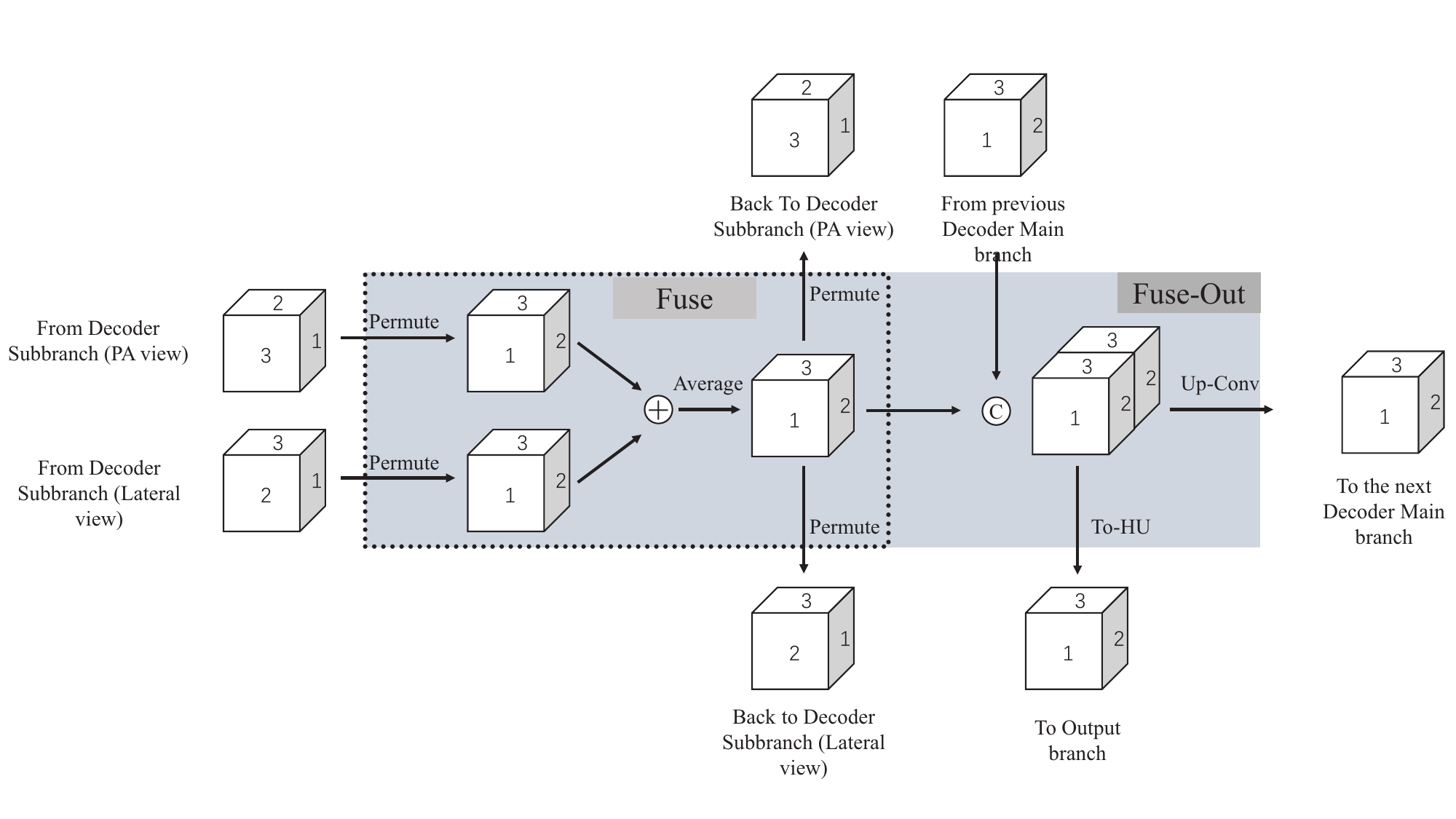}
    \caption{Illustration of the basic fusion block of MFusionRend Module: Fuse and Fuse-out block. They receive information from both decoders and perform information fusion. The Fuse-out block is an enhanced version of the Fuse block with an additional information output function.}
    \label{fig:MSFR}
\end{figure}

X-Recon involves a dimensional conversion between 2D and 3D, necessitating specific transformations of the feature map. To expand 2D feature maps into their corresponding 3D counterparts, they are duplicated and stacked along the channel dimension in the skip connection. Additionally, a transformation of the neuron numbers through fully connected layers, as shown in the detailed view of the 2D-3D block in \Cref{fig:generator}, is necessary between the encoder and decoder to accommodate the dimensions required for the rearranged 3D feature map. Therefore, X-Recon incorporates a multi-scale fusion rendering (MFusionRend) module to produce the final CT reconstruction results. As outlined in \Cref{sec:generator}, the features extracted from the two views by the twin encoder-decoder networks are subsequently unified within the fusion and decoding module (\Cref{fig:MSFR}). The average feature map is subsequently calculated and transmitted back to the two decoder branches, facilitating the fusion of information from a dual orthogonal view. During the fusion of ortho-lateral X-ray features, the rendering module is employed to predict CT values for feature maps of various scales. CT values of different sizes are normalized and superimposed through up-sampling to achieve multi-scale rendering, ranging from coarse-grained to fine-textured, in the output branch.

\subsection{PTX-Seg: Zero-shot Pneumothorax Segmentation Framework}\label{Sec:PtxSeg}
For pneumothorax segmentation, we introduced a zero-shot segmentation framework called PTX-Seg, which combines learning-based methods with traditional image processing techniques based on the anatomical priors. PTX-Seg is a two-branch segmentation framework with detailed procedures outlined in the \Cref{algo:PTX-Seg}. In the learning-based branch, we initially perform lung parenchyma segmentation using the widely recognized deep learning method, specifically, U-Net, with pre-trained weights obtained from the winning solution of the LOLA11 Challenge~\cite{hofmanninger_automatic_2020}. The input CT image $I$ will undergo multiple frozen U-Net $\mathcal{F}_{i}(\cdot)$ to obtain lung parenchyma regions $M_{l}$. For robust segmentation, the union model ensemble is utilized to ensure that the lung parenchyma is contained in $M_{l}$ and followed by a closed operation with kernel $K_{l}$.

Since pneumothorax is not included in the training set of U-Nets, we require the auxiliary branch to obtain air-accumulated pneumothorax regions that incorporate anatomical priors. This branch begins by binarizing the entire CT sequence to provide an initial separation of the patient's body and regions containing air. After the threshold binarization $\operatorname{Threshold(\cdot)}$ with the intensity threshold $t_{b}$, the coarse body mask $M_{bc}$ is obtained. However, it's worth noting that the air is also present in the trachea and lungs, which could be erroneously categorized as part of the body mask. To address this, a 3D closure operation using kernel $K_{b}$ is applied to fill small holes within the lungs and close off the tracheal region. Subsequently, the mask of the non-body regions $M_{b}^{\prime}$ could be obtained using seed-point-based region-growing operation $\operatorname{Grow}(\cdot)$ with growing pattern $T$ using two corner points, i.e. $\boldsymbol{x}_{0}=(0,0,0)$ and $\boldsymbol{x}_{max} = (D, W, H)$. Thus, inverting $M_{b}^{\prime}$ yields the fine body mask $M_{b}$. The coarse segmentation of pneumothorax region $M_{pc}$ could be obtained by using the intersection of lung parenchyma regions $M_{l}$ from the learning branch and the body region from $M_{b}$, followed by the binarization with a threshold value of $t_{p}$. Finally, the segmentation mask of pneumothorax $M_{p}$ is achieved by refining $M_{pc}$ by identifying the portion with CT values less than -950 (as the CT value of air is -1000) and subsequently using a 3D closure operation is applied to the extracted area, and regions with air volume threshold $v_{t}$ exceeding 10 ml are selected as the final pneumothorax air accumulation regions.

\IncMargin{1em}
\begin{algorithm}[tbp]
\begin{small}
\SetKwInOut{Input}{Input}\SetKwInOut{Output}{Output}
\SetKwFor{For}{For}{}{End for}
\SetKw{KwRet}{Return}
\leftskip=-1.2em
\Input{CT image $I$.}
\Output{Fine-grained air-accumulated pneumothorax region $M_{p}$ of $I$.}
\leftskip=0em
\BlankLine
\BlankLine
\tcc*[h]{Segmentation of lung parenchyma}\;
\nl$M_{l} \leftarrow \mathbf{0} \in \mathbb{R}^{D \times W \times H}$\;

\nl\For{$i$ in $1, 2,\ldots,$ U-Net Nums}
{
    \tcc*[h]{get unions of multiple U-Nets}\;
    $M_{l} \leftarrow M_{l} \bigcup \mathcal{F}_{i}(I)$\;
}
\nl$M_{l} \leftarrow M_{l} \circ K_{l}$\;
\tcc*[h]{Auxiliary branch}\;
\nl$M_{bc} \leftarrow \operatorname{Threshold}(I, t_{b}) \circ K_{b}$\;
\nl$M_{b} \leftarrow (\operatorname{Grow}(M_{bc}, \boldsymbol{x}_{0}, T) \bigcup \operatorname{Grow}(M_{bc}, \boldsymbol{x}_{max}, T) )^{\prime} $\;
\tcc*[h]{Mask ensembling and obtain air-accumulated region}\;
\nl$M_{pc} \leftarrow \operatorname{Threshold}(M_{l} \bigcap M_{b}, t_{p})$\;
\nl$M_{p} \leftarrow \operatorname{Measure}(M_{pc} \circ K_{p}, v_{t})$\;
\nl\KwRet{$M_{p}$}
\end{small}
\caption{Pipeline of PTX-Seg}\label{algo:PTX-Seg}
\end{algorithm}\DecMargin{1em}

\subsection{Loss functions of X-Recon}
\subsubsection{Reconstruction loss}
In the context of CT reconstruction, X-Recon aims to ensure a high level of structural consistency between the reconstructed CT and the real CT acquired from patients. As a result, constraints are necessary to enhance the consistency between the reconstructed CT and the real CT at the voxel level.

As CT is a 3D single-channel image, which demands a high level of structural accuracy within objects, this study employs the L2 loss as a constraint for reconstruction in X-Recon. This loss is used to attain reconstruction results that closely resemble the real CT structure of the patient. The reconstruction loss is defined as follows:
\begin{equation}
    \mathcal{L}_{Re}=\mathbb{E}(||y-\mathcal{G}(x)||_2)
\end{equation}
where $y$ represents the ground truth CT image, and $\mathcal{G}(x)$ represents the CT image reconstructed by the generator $\mathcal{G}(\cdot)$.

\subsubsection{Digitally Reconstructed Radiograph loss (DRR loss)}
In addition to the utilization of L2 loss as a constraint in 3D space, X-Recon also seeks to impose constraints on the overall structure in 2D projection space. Inspired by image registration techniques~\cite{gao_generalizing_nodate} that align X-ray and CT images, X-Recon generalizes the spatial transformer to projection geometry, introducing what is referred to as the Projective-Spatial-Transformer (ProST), which enables differentiable projection rendering. The ProST facilitates end-to-end image processing and gradient-based optimization in X-Recon. Another objective of X-Recon is to ensure the consistency of the 2D projection of the reconstructed CT with the projection of the real CT from various image views. Therefore, X-Recon follows conventions in medical imaging, opts to constrain the projections in three orthogonal views: Posteroanterior, Axial, and Lateral. To enhance robustness, X-Recon introduces stochastic viewpoint rotation and translation during training. Similar to the classical Pix2pix model~\cite{isola_image--image_nodate}, X-Recon employs L1 loss to ensure the output results align with the input conditions. In this context, L1 loss is applied to multi-angle projected images to enhance image edge sharpness. This defines a multi-angle projection loss, expressed as follows:
\begin{align}
\begin{aligned}
\mathcal{L}_{DRR}=&\frac{1}{3}(\mathcal{L}_{DRR_{Pa}}+\mathcal{L}_{DRR_{La}}+\mathcal{L}_{DRR_{Ax}})\\
            =&\frac{1}{3}[\mathbb{E}(||\mathcal{P}_{Pa}(y)-\mathcal{P}_{Pa}(\mathcal{G}(x))||_1)+\mathbb{E}(||\mathcal{P}_{Ax}(y)\\
            &-\mathcal{P}_{Ax}(\mathcal{G}(x))||_1) +\mathbb{E}(||\mathcal{P}_{La}(y)-\mathcal{P}_{La}(\mathcal{G}(x))||_1)]
\end{aligned}
\end{align}
where PA represents the posteroanterior view, Ax stands for the axial view, and La refers to the lateral view. This formula computes the sum of the L1 losses of the projected views between the real CT and the reconstructed CT for all views.

\subsubsection{Generative adversarial loss (GAN loss)}
The generative adversarial network (GAN)~\cite{goodfellow_generative_2020} is a classical generative model renowned for its capability to formulate real data distributions. The conditional generative adversarial network (cGAN)~\cite{mirza_conditional_2014} is one of the extensions of the original GAN, specifically designed to supervise and enhance the data generation process. 
The supervised generation is achieved by using additional information as conditional inputs to the model which can be in the form of category labels, segmentation masks, and even data for different modalities.
Inspired by cGAN, X-Recon introduces prior information, namely 2D X-ray images, into its training process. Given that the reconstruction task involves a transformation from X-ray to CT, necessitating the preservation of the same semantic information as the original X-ray, X-Recon incorporates the 2D X-ray image as prior knowledge within the generator to guide the CT reconstruction process. Furthermore, this prior condition is also integrated into the discriminator to elevate its learning capabilities, thereby motivating the generator to enhance the quality of the reconstructed CT images. The GAN model is trained using the least squares generative adversarial network (LSGAN) loss, and the loss is defined as follows:
\begin{align}
\begin{aligned}
    \mathcal{L}_{D}=&\frac{1}{2}[\mathbb{E}_{y \sim p_{CT}, x \sim p_{XRay}}(||\mathcal{D}(y|x)-1||_{2})\\
                         &+\mathbb{E}_{x \sim p_{XRay}}(||\mathcal{D}(\mathcal{G}(x)|x)-0||_{2})]
\end{aligned}
\end{align}
\begin{equation}
    \mathcal{L}_{G}=\frac{1}{2}[\mathbb{E}_{x \sim p_{XRay}}(||\mathcal{D}(\mathcal{G}(x)|x)-1||_{2})]
\end{equation}
where $\mathcal{D}(y|x)$ represents the discriminator's assessment of the real CT providing prior conditions of input 2D chest X-ray images, and $\mathcal{D}(\mathcal{G}(x)|x)$ stands for the discriminator's evaluation of the CT reconstructed by the generator under the same prior condition.

\section{Experiments}
\subsection{Dataset collection}
This study collected data from a total of 255 patients with pneumothorax between January 2015 and October 2021. Additionally, a control group consisting of 279 healthy subjects with normal CT images from July 2019 to October 2021 was also included. In total, this study included CT sequences from 534 subjects. The data collection, usage, and research protocol are approved by the Ethics Committee of the Shanghai Public Health Clinical Centre. The Picture Archiving and Communication System (PACS) of the Public Health Clinical Centre was utilized to identify patients with spontaneous or traumatic pneumothorax. To ensure stable performance analysis and fair performance comparison, collected images undergo visual review by clinical experts, and low-quality images are excluded from this study. After data standardization, each patient corresponds to only one CT examination. All available chest CT images were collected, and the dataset was anonymized by removing patient privacy-related records. For the performance validation and analysis of PTX-Seg, a senior radiologist was invited to annotate and delineate both lungs and regions of pneumothorax using the ITK-SNAP tool as the gold standard in the subsequent segmentation evaluation. 

Considering the practical challenge of acquiring paired X-rays and CTs due to the high cost and ethical concerns associated with exposing patients to additional radiation, this study employed a Digitally Reconstructed Radiograph (DRR)~\cite{chaney_portable_1995} technique for data simulation. The DRR was used to train the X-Recon network using patients' CT data, generating X-ray images by simulating an X-ray source and detector. More specifically, when provided with a CT sequence, the DRR technique was utilized to generate simulated chest X-ray images in both the posteroanterior view (PA view) and lateral view, effectively producing ortho-lateral chest X-ray images.

\subsection{Implementation Details}
\textbf{Training Pipeline:}
The training and test sets were divided according to a four-to-one ratio. The Network was trained using the Adam optimizer with an initial learning rate of $5e-5$, betas = (0.5, 0.99), and weight decay of $5e-4$ with a linear decay of the learning rate to 0 in the subsequent 100 rounds of training. The input X-ray image size was set to 224 $\times$ 224, and the output CT image size was set to 224 $\times$ 224 $\times$ 224. To generate high-resolution and high-accuracy CTs in an end-to-end manner within the constraints of limited graphics processing unit (GPU) memory, the batch size is set to 1 in our implementation. Under conditions with small batch size, X-Recon utilized instance normalization, and the process is accelerated by an Nvidia A100 GPU with 40 GB video random access memory (VRAM).

\subsection{Segmentation Performance of PTX-Seg}
Firstly, the segmentation performance of PTX-Seg was quantitatively evaluated (e.g., \Cref{tab:seg}), using the annotations labeled by radiologists as the ground truth to assess the segmentation accuracy for both lungs as well as regions. A total of four metrics were used to assess the segmentation performance, i.e. Dice coefficient, Jaccard coefficient, 95th percentile of Hausdorff Distance ($HD_{95}$), and Average Surface Distance (ASD). As can be seen from \Cref{tab:seg}, the segmentation accuracy of PTX-Seg for the right lung, left lung, and air-accumulated regions is $96.93\%$, $97.54\%$, and $96.32\%$, respectively, demonstrating excellent segmentation precision for these objects.

\begin{table}[htbp]
    \vspace{-0.2 cm}
    \centering
    \caption{Segmentation accuracy assessment of the PTX-Seg, a quantitative measure of its segmentation performance for the lung and air regions}
    \resizebox{\linewidth}{!}{
        \begin{tabular}{ccccc}
        \toprule
                & $Dice (\%)$ & $Jaccard (\%)$ & $HD_{95}(mm)$  & $ASD(mm)$ \\
        \midrule
        Right lung & 96.93 & 94.24 & 9.52  & 0.12 \\
        Left lung & 97.54 & 95.40  & 7.61  & 0.10 \\
        Air regions & 96.32 & 93.27 & 1.35  & 0.29 \\
        \bottomrule
        \end{tabular}%
    }
    \label{tab:seg}%
    \vspace{-0.6cm}
\end{table}%

\subsection{Reconstruction results of X-Recon}
\subsubsection{Qualitative analysis}
\begin{figure}[htbp]
    \vspace{-0.2 cm}
    \centering
    \includegraphics[width=\linewidth]{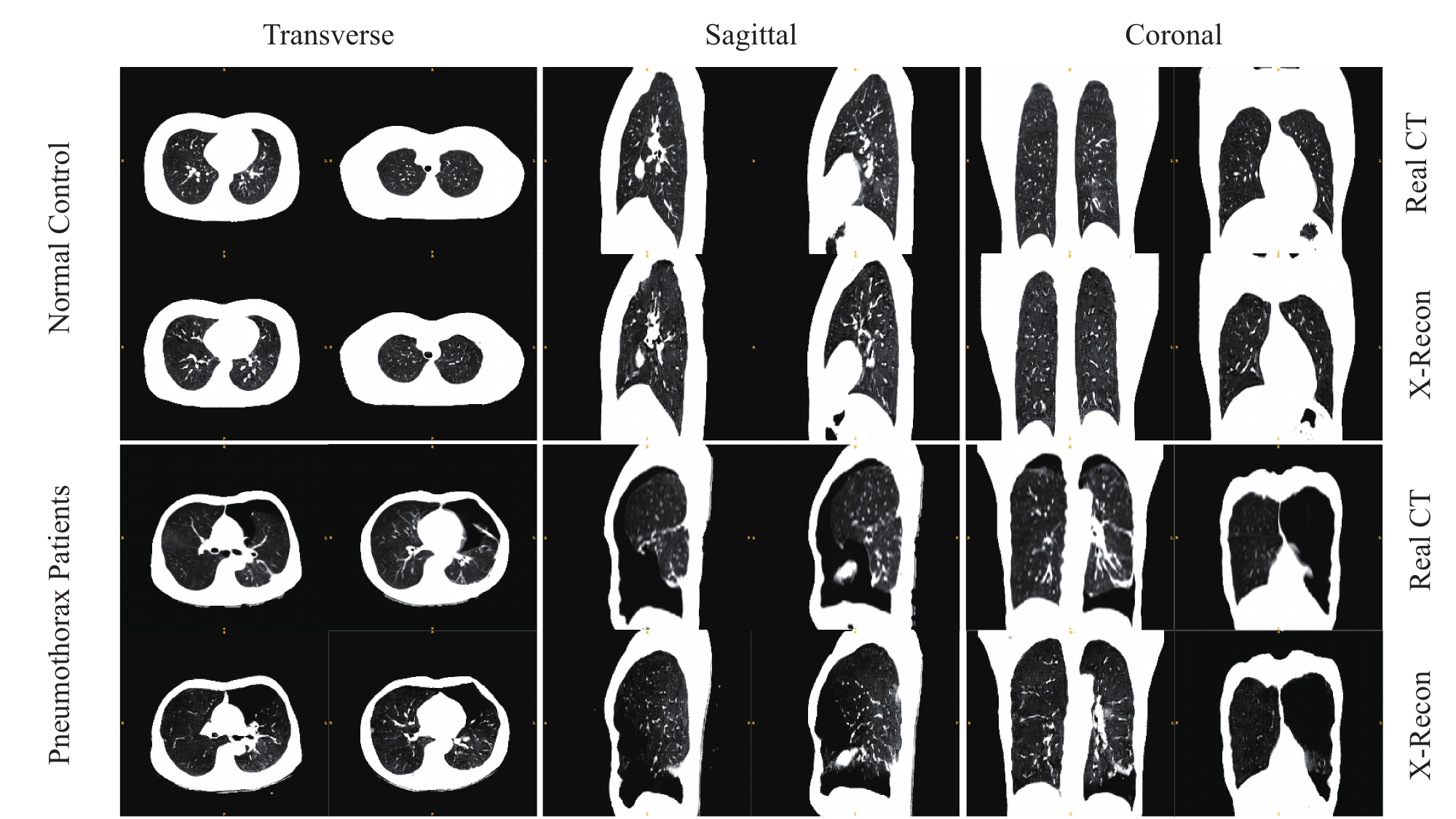}
    \caption{Visualization of X-Recon reconstruction results. The first two rows are samples from healthy subjects and the last two rows are samples from pneumothorax patients. The images are presented in three anatomical planes: transverse, sagittal, and coronal planes.}
    \label{fig:vis}
    \vspace{-0.6 cm}
\end{figure}

:
The reconstruction results of X-Recon are visualized in \Cref{fig:vis}, using examples of a healthy individual and a patient with pneumothorax. Real CT images and X-Recon reconstructed images are displayed from three anatomical perspectives: transverse, sagittal, and coronal views. In the case of the healthy individual, both lung parenchyma are clearly visible in the reconstructed CT, and the position and morphology of each organ match those in the real CT. In patients with pneumothorax, CT images show rims of gas around the edges of the lung which may track up the fissures, and compressed lung edges are visible at the inner margins. In the reconstructed CT images of pneumothorax patients, the tracheal direction closely resembles that of the real CT, the pneumothorax area and lung boundaries are distinctly discernible, and the volume and location of the pneumothorax tend to match those in the real CT.

\subsubsection{Comparison with reconstruction methods}
For qualitative and quantitative evaluation of reconstruction results, we compared our proposed X-Recon with SOTA methods in this field. Since few published works focus on solving the X-ray to CT ultra-sparse reconstruction problem, this study reproduced two learning-based single-view reconstruction methods~\cite{shen_patient-specific_2019,henzler_singleimage_nodate} for comparison.

\begin{figure}[htbp]
    \centering
    \vspace{-0.2 cm}
    \includegraphics[width=\linewidth]{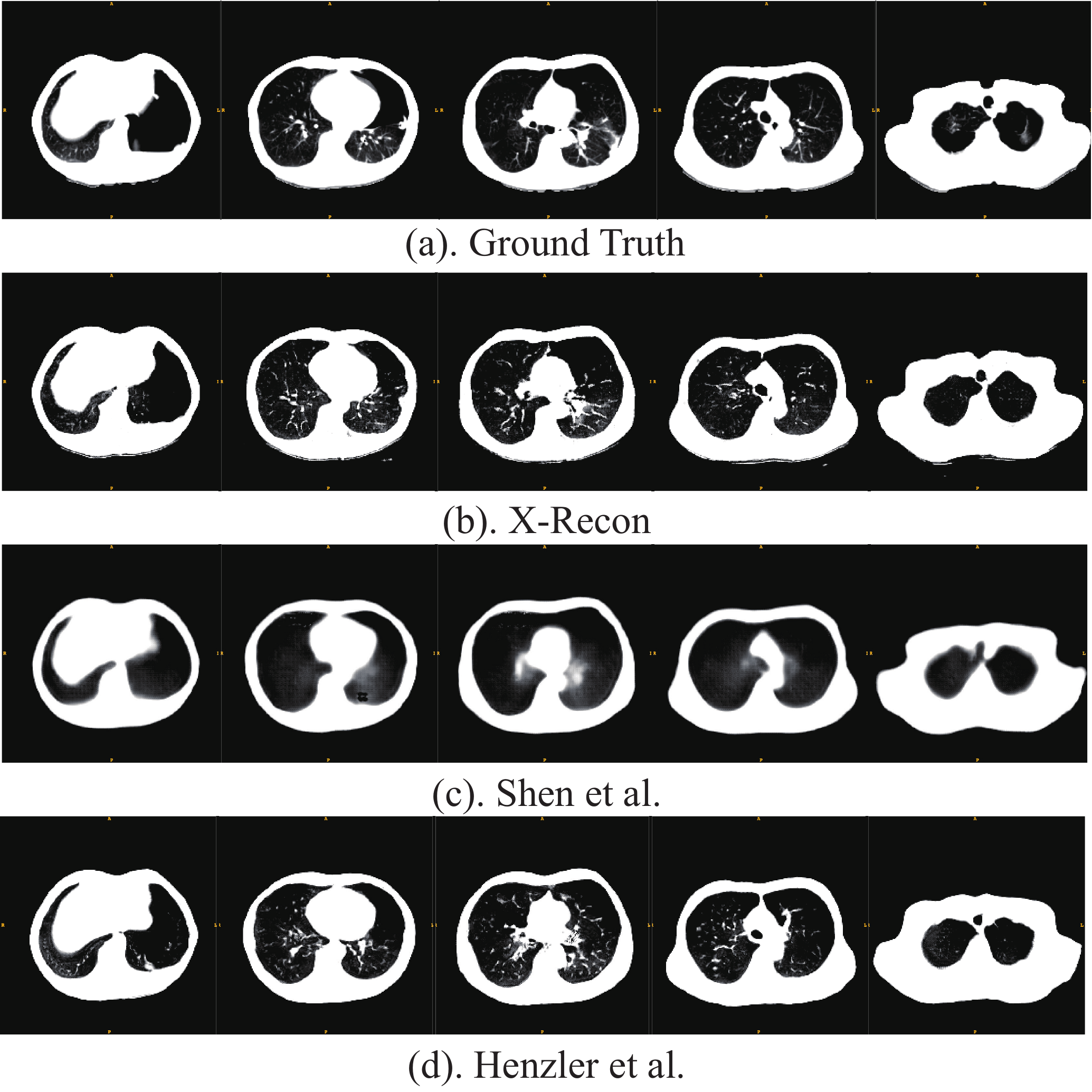}
    \caption{Qualitative comparison with reconstruction methods: The first row shows the real CT image of a typical pneumothorax patient, the second row is reconstruction results of X-Recon, and the third and fourth rows are reconstructions using the single-view sparse reconstruction methods proposed by Shen et al., and Henzler et al.}
    \label{fig:sota}
    \vspace{-0.2 cm}
\end{figure}

\begin{figure}[tbp]
    \vspace{-0.8cm}
    \centering
    \includegraphics[width=\linewidth]{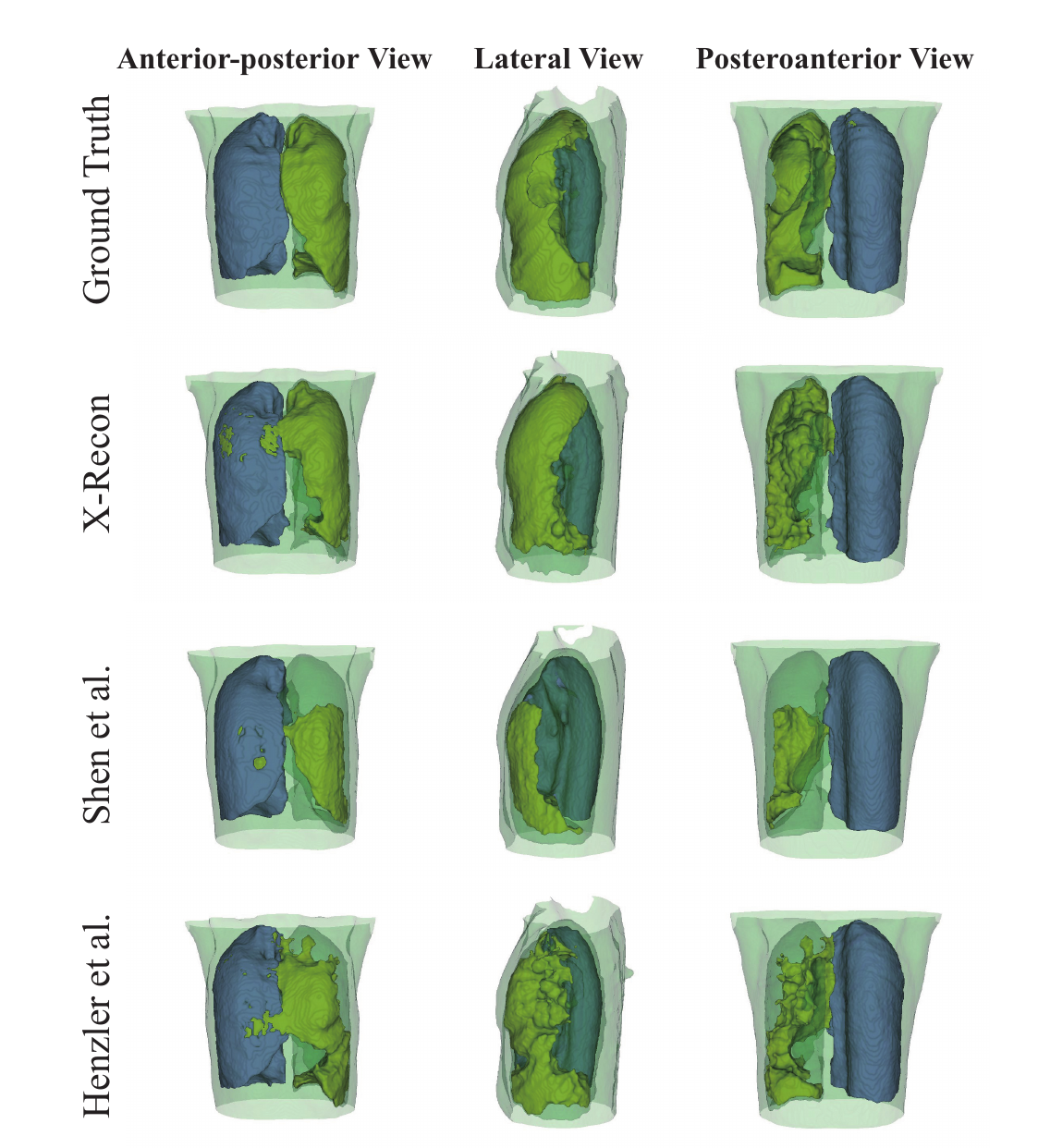}
    \caption{Comparison with other methods by 3D rendering. The yellow section indicates the area of air accumulation and the blue section is the right lung. The first row is the 3D rendering of both lungs and the area of air accumulation in a typical pneumothorax patient, the second row X-Recon is our proposed dual-view reconstruction method, and the third and fourth rows are the single-view sparse reconstruction methods proposed by Shen et al. and Henzler et al.}
    \label{fig:vis3d}
    \vspace{-0.2cm}
\end{figure}

From a qualitative perspective, this study selected a typical CT scan of a patient with pneumothorax for visualization to compare the effectiveness of different reconstruction methods. The methods proposed by Shen et al.~\cite{shen_patient-specific_2019} and Henzler et al.~\cite{henzler_singleimage_nodate} are designed for single-view X-ray images and lack the information provided by other X-ray angles, as shown in \Cref{fig:sota}. The visual quality assessment in this study reveals a distinct difference. Shen et al.'s single-view reconstruction method produces a blurred CT, losing details of the lungs despite relatively good morphology for large organs like the heart and chest wall. Henzler et al.'s method simulates lung parenchyma reasonably well but faces challenges in reconstructing the air-accumulated region. In contrast, X-Recon 
excels in capturing fine-grained structural changes in the lungs and provides a superior reconstruction of the air-accumulated region in patients with pneumothorax. Furthermore, in \Cref{fig:vis3d}, the segmentation results based on PTX-Seg for real CT and CT images generated by different reconstruction methods are displayed in 3D rendering, visualized from anterior-posterior, lateral, and posteroanterior views. These results demonstrate that the CT reconstructed by X-Recon exhibits a high degree of similarity to the real CT in both the left and right lungs, as well as in the air-accumulated region.

From a quantitative perspective, this study also conducts evaluations based on image reconstruction quality, pneumothorax diagnostic performance, and image segmentation metrics. The image reconstruction quality was evaluated on three metrics, including Cosine Similarity (CS), Peak Signal-to-Noise Ratio (PSNR), and Structural Similarity (SSIM). Additionally, this study assessed the diagnostic value of X-Recon's reconstructed CT for pneumothorax. The zero-shot pneumothorax segmentation algorithm PTX-Seg served as the silver standard and was applied to extract the right and left lung parenchyma and air-accumulated regions from the real patient CT and reconstructed CTs. Volumes and pleural cavity occupancy ratios of the air-accumulated region were calculated. Pearson correlation coefficients were used to measure the similarity between the real CT and reconstructed CT in terms of volume and pneumothorax occupancy. The evaluation results are presented in \Cref{tab:sota}. Besides, the segmentation results for both the lungs and air-accumulated regions were assessed using metrics such as Dice coefficient, Jaccard's coefficient, Hausdorff's distance, and mean surface distance. Overall, as shown in \Cref{tab:app}, X-Recon outperforms other methods in terms of image generation, pneumothorax diagnosis, and segmentation of critical regions, demonstrating superior performance.

\begin{table}[t]
    \vspace{-0.6 cm}
    \centering
    \caption{Quantitative comparison with other methods using metrics related to image reconstruction quality, pneumothorax diagnosis, and image segmentation, respectively. CS: cosine similarity, PSNR: peak signal-to-noise ratio, SSIM: structural similarity, CC: the correlation coefficient between generated image and ground-truth CT, RLCC: CC of the right lung volume, LLCC: CC of the left lung volume, ARCC: CC of air accumulated region volume, OCC: CC of the pleural cavity occupancy ratios of the air-accumulated region. Bold indicates the best performance.}
    \resizebox{\linewidth}{!}{
        \begin{tabular}{cccc}
        \toprule
                & Henzler et al. & Shen et al. & X-Recon \\
        \midrule
        CS    & 0.81  & 0.81  & \textbf{0.89} \\
        PSNR  & 17.36 & 17.35 & \textbf{19.85} \\
        SSIM  & 0.64  & 0.66  & \textbf{0.70} \\
        RLCC  & 0.77  & 0.65  & \textbf{0.84} \\
        LLCC  & 0.82  & 0.70   & \textbf{0.89} \\
        ARCC& 0.69  & 0.22  & \textbf{0.82} \\
        OCC& 0.62  & 0.21  & \textbf{0.77} \\
        \bottomrule
        \end{tabular}%
    }
    \label{tab:sota}%
\end{table}%

\begin{table}[htbp]
    \centering
    \caption{Quantitative comparison of X-Recon with other methods by metrics related to image segmentation, including Dice coefficient, Jaccard coefficient, $95_{th}$ percentile Hausdorff Distance ($HD_{95}$), and average surface distance ($ASD$). Bold indicates the best performance.}
    \vspace{-0.2cm}
    \resizebox{\linewidth}{!}{
    \begin{tabular}{cccccc}
    \toprule
    Methods & Partition & Dice (\%) & Jaccard (\%) & $HD_{95}\ (mm)$  & $ASD\ (mm)$ \\
    \midrule
    \multirow{3}[1]{*}{Henzler et al.} & Right Lung & 87.57 & 79.65 & 7.10   & 2.32 \\
          & Left Lung & 84.92 & 76.02 & 7.24  & 2.55 \\
          & Air Region & 57.14 & 54.55 & 77.76 & 68.10 \\
    \multirow{3}[0]{*}{Shen et al.} & Right Lung & 83.74 & 74.58 & 9.91  & 2.86 \\
          & Left Lung & 83.24 & 74.55 & 8.81  & 2.50 \\
          & Air Region & 16.48 & 15.29 & 150.44 & 136.02 \\
    \multirow{3}[1]{*}{X-Recon} & Right Lung & \textbf{92.07} & \textbf{86.48} & \textbf{4.32}  & \textbf{1.34} \\
          & Left Lung & \textbf{90.21} & \textbf{83.88} & \textbf{4.76}  & \textbf{1.57} \\
          & Air Region & \textbf{63.83} & \textbf{60.99} & \textbf{63.44} & \textbf{53.74} \\
    \bottomrule
    \end{tabular}%
    }
    \label{tab:app}%
    \vspace{-0.4 cm}
\end{table}%

\subsection{Ablation Experiments}
Subsequently, this study conducted ablation experiments on X-Recon, focusing on network structure and loss functions.
\subsubsection{Ablation experiments on the network structure}
This study conducted ablation experiments on the coordinate convolution in the discriminator network, the skip connection in the generator network, and the multi-scale fusion module, respectively. As presented in Table.~\ref{tab:abl_net}, X-Recon, as proposed in this study, achieved superior results with a cosine similarity of 0.89, a peak signal-to-noise ratio of 19.85, and a structural similarity of 0.70. These results demonstrate the effectiveness of the proposed method, outperforming the other compared approaches.

\begin{table}[hbtp]
    \vspace{-0.6cm}
    \centering
    \caption{Ablation experiments on the network structure of X-Recon, using metrics related to image reconstruction. \checkmark represents the result with that setting, while blank represents the result without that setting.}
      \resizebox{\linewidth}{!}{
        \begin{tabular}{cccccc}
        \toprule
        \multicolumn{1}{p{5.415em}}{Coordinate Convolution} & \multicolumn{1}{p{5.96em}}{Skip Connection} & \multicolumn{1}{p{5em}}{Multiscale Fusion}   & CS    & PSNR  & SSIM \\
        \midrule
            & \checkmark    & \checkmark    & 0.89  & 19.76 & 0.69 \\
        \checkmark    &     & \checkmark    & 0.86  & 18.57 & 0.67 \\
        \checkmark    & \checkmark    &     & 0.87  & 18.72 & 0.67 \\
        \checkmark    &     &     & 0.86  & 18.62 & 0.67 \\
        \checkmark & \checkmark & \checkmark & \textbf{0.89} & \textbf{19.85} & \textbf{0.70} \\
        \bottomrule
        \end{tabular}%
    }
    \label{tab:abl_net}%
\end{table}%

\subsubsection{Ablation experiments with loss functions}
Ablation experiments were conducted to evaluate the impact of different loss functions on X-Recon, including reconstruction loss, multi-angle projection loss, and adversarial loss. Both qualitative and quantitative assessments were performed to provide an accurate and comprehensive evaluation of the loss functions.

\begin{figure}[htbp]
    \vspace{-0.2cm}
    \centering
    \includegraphics[width=\linewidth]{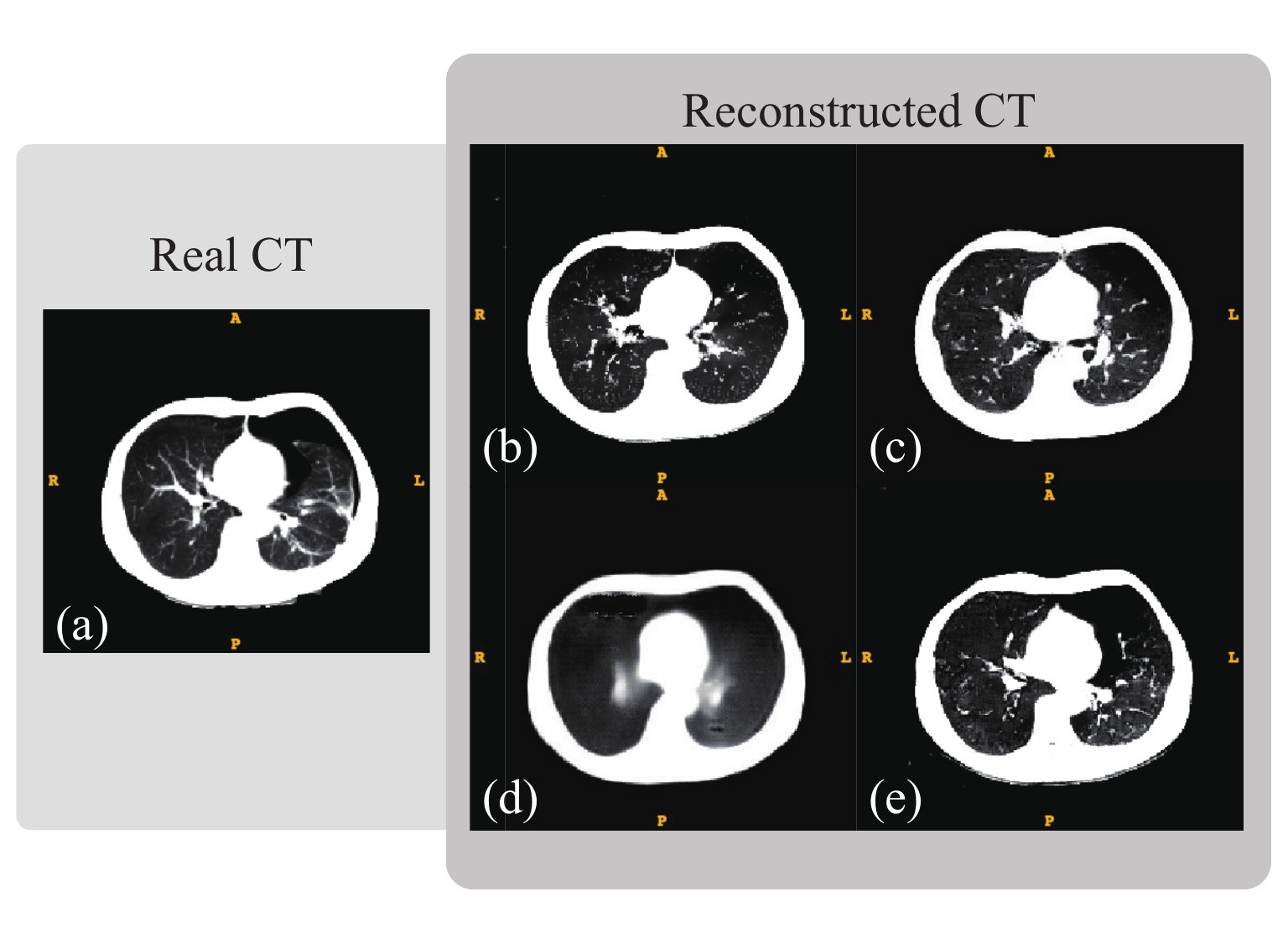}
    \caption{Visualization examples of X-Recon loss function ablation experiments. (a) Real CT image; (b) Reconstruction results without reconstruction loss; (c) Reconstruction results without digitally reconstructed radiograph loss; (d) Reconstruction results without generative adversarial loss; (e) Reconstruction results of our proposed X-Recon}
    \label{fig:loss}
    \vspace{-0.2cm}
\end{figure}

From a qualitative perspective, the visualizations of X-Recon trained with different loss functions are presented in \Cref{fig:loss}. When the generative adversarial loss was removed (\Cref{fig:loss} (d)), the lung parenchyma in the reconstructed CT appeared blurred. If either the reconstruction loss or multi-angle projection loss was removed (\Cref{fig:loss} (b)-(c)), the lung parenchyma improved, but the boundary between the air-accumulated region and the lung parenchyma remained unclear, leading to difficulties in calculating pneumothorax volume. In contrast, X-Recon trained with all three loss functions, namely reconstruction loss, multi-angle projection loss, and generative adversarial loss, produced the best reconstruction results. The reconstructed CT had clear lung parenchyma and a well-defined boundary between the air-accumulated region and lung parenchyma (\Cref{fig:loss} (e)), outperforming any other configurations.

From a quantitative perspective, the results presented in \Cref{tab:abl_loss} demonstrated that X-Recon using all proposed loss functions achieved the highest performance for each evaluation metric. The correlation coefficients for the volumes of the left lung, right lung, and air-accumulated region were all above 0.8, with a high correlation coefficient of 0.77 for pleural cavity occupancy. This highlights the clinical diagnostic potential of the reconstructed CT produced by X-Recon trained with all three specified loss functions.

\begin{table}[htbp]
    \centering
    \caption{Ablation experiments on the loss function of X-Recon, using metrics related to pneumothorax diagnosis. The last 3 columns are the generated proportion of different structures. CC: the correlation coefficient for the volumes between the generated image and ground-truth CT, RLCC: CC of the right lung, LLCC: CC of the left lung, ARCC: CC of air accumulated region, OCC: CC of the thoracic cavity occupancy ratios of the air-accumulated region. \checkmark represents the result with that loss, while blank represents the result without that loss.}
    \resizebox{\linewidth}{!}{
        \begin{tabular}{ccccccc}
        \toprule
        \multicolumn{1}{c}{\textbf{$\mathcal{L}_{Re}$}} & \multicolumn{1}{p{4.04em}}{\textbf{$\mathcal{L}_{DRR}$}} & \multicolumn{1}{p{4.04em}}{\textbf{$\mathcal{L}_{D}+\mathcal{L}_{G}$}} & RLCC& LLCC& ARCC& OCC\\
        \midrule
            & \checkmark     & \checkmark     & 0.72  & 0.77  & 0.64  & 0.53 \\
        \checkmark     &     & \checkmark     & 0.77  & 0.84  & 0.72  & 0.63 \\
        \checkmark     & \checkmark     &     & 0.76  & 0.79  & 0.55  & 0.52 \\
        \textbf{\checkmark} & \textbf{\checkmark} & \textbf{\checkmark} & \textbf{0.84} & \textbf{0.89} & \textbf{0.82} & \textbf{0.77} \\
        \bottomrule
        \end{tabular}%
    }
    \label{tab:abl_loss}%
    \vspace{-0.4cm}
\end{table}%

\section{Discussion}

This study introduced a novel learning-based method for reconstructing CT images using biplane X-rays acquired from orthogonal views, referred to as X-Recon. The performance of X-Recon is rigorously evaluated through both qualitative and quantitative experiments. Additionally, a zero-shot pneumothorax segmentation algorithm with high segmentation precision, namely PTX-Seg, is proposed and used for further quantification evaluation for image reconstruction. The PTX-Seg could reach an excellent segmentation accuracy in terms of Dice, which obtained 96.93\%, 97.54\%, and 96.32\% for the right lung, left lung, and air-accumulated region respectively. As shown in \Cref{fig:vis}, the dual-view reconstruction method X-Recon realized a high-resolution ultra-sparse reconstruction performance in reconstructing the structure of large organs such as the heart, chest wall and maintaining detailed information of the lung parenchyma while simulating the diseased areas. The high-quality reconstructed CT demonstrated the feasibility of using ortho-lateral X-ray reconstruction for chest disease diagnosis. In the subsequent quantitative experiments, we evaluated the effectiveness of different network modules and the influence of different loss functions on X-Recon. 

Furthermore, X-Recon was employed in the diagnostic task of pneumothorax and used for quantitative evaluation of clinical metrics. As the most clinically interesting and practically valuable metric is the percentage of the pleural cavity occupied by the pneumothorax air-accumulated region, this study used the proposed pneumothorax segmentation algorithm PTX-Seg to segment and statistically measure the relevant regions in both the patient's real CT and the reconstructed CT. Besides, this study obtained the percentage of pleural cavity occupied by pneumothorax in both the real CT and the reconstructed CT. The results show that Pearson's correlation coefficient for this indicator reached 0.77, and Pearson's correlation coefficients for indicators such as lung parenchyma and air-accumulated volume exceeded a high correlation of 0.8. This indicates that the CT reconstructed by X-Recon holds promise for clinical applications.

Although X-Recon performs well in multiple experiments, it does have some limitations to be improved in future studies. Considering the practical challenges of obtaining paired X-rays and CTs due to cost and ethical concerns related to additional patient radiation exposure, this study used X-rays simulated by the DRR technique instead of real patient scans. Although the DRR-simulated X-rays exhibit good realism, there is still room for improvement in capturing fine tissue structures like the trachea and blood vessels. Additionally, all CT scans were acquired in the supine position, resulting in the chest X-rays synthesized by DRR also being in the supine position. This differs from the target clinical scenario of standing position, which may impact the performance of clinical applications. Additionally, there is still room for improving the resolution of the reconstructed CT. Although the current reconstructed CT of X-Recon has achieved a resolution of 224 $\times$ 224 $\times$ 224, which is already higher than the current SOTA methods with resolutions of 128 $\times$ 128 $\times$ 128~\cite{shen_patient-specific_2019,henzler_singleimage_nodate}, its spatial resolution, averaging 1 to 2 mm, still falls short of that offered by high-resolution thin-layer CT scans.

Therefore, in future works, it's essential to address the problem of the discrepancy between DRR-simulated X-rays and real X-rays. In the realm of algorithms, future developments are expected to be supplemented by research in style migration to reduce such discrepancies. In terms of data, studies on the phantom may offer solutions to overcome data acquisition limitations and produce paired real chest X-ray and CT data. To achieve a higher resolution of reconstructed CT, novel methodologies and theories can be derived and introduced from the recent advances of the vision community to simplify the network architecture and further increase the resolution of the reconstructed CT.

\section{Conclusion}
This study proposed a learning-based X-Recon network, designed to reconstruct high-resolution CT images based on ortho-lateral chest X-rays. Additionally, a zero-shot pneumothorax segmentation algorithm, PTX-Seg, is proposed for patient-specific correlation analysis. The expanded experiments demonstrate that X-Recon can produce high-precision CT sequences with reduced radiation exposure and cost. Furthermore, the study enables accurate three-dimensional quantitative assessment of pneumothorax-related indicators. X-Recon's capabilities can also be extended to diagnose other chest-related diseases, potentially inspiring more researchers to address this challenge and advance the field of precision medicine.


\balance
\bibliography{IEEEabrv, bibtex}
\bibliographystyle{IEEEtran.bst}
\end{document}